\documentclass{article}
\usepackage{spconf,amsmath,multirow}
\usepackage[pdftex]{graphicx}
\graphicspath{{./figs/}}
\usepackage[caption=false,font=footnotesize]{subfig}
\usepackage{hyperref}
\usepackage{arabtex}
\usepackage{utf8}
\setcode{utf8}
\begin{document}

\title{AlexU-Word: A New Dataset for Isolated-Word Closed-Vocabulary Offline Arabic Handwriting Recognition}

\name{Mohamed E. Hussein\sthanks{Mohamed E. Hussein is currently an Assistant Professor at Egypt-Japan University of Science and Technology, New Borg El-Arab City, Alexandria, Egypt; on leave from his position at Alexandria University, where most of the work associated with this project was performed.}, Marwan Torki, Ahmed Elsallamy, and Mahmoud Fayyaz}
\address{Computer and Systems Engineering Department, \\
Faculty of Engineering, Alexandria University\\
mehussein@alexu.edu.eg, mtorki@alexu.edu.eg, sallamy1991@gmail.com, fayyaz.method@gmail.com}

\maketitle

\begin{abstract}
% The abstract goes here. DO NOT USE SPECIAL CHARACTERS, SYMBOLS, OR MATH IN YOUR TITLE OR ABSTRACT.
In this paper, we introduce a new dataset for offline Arabic handwriting recognition. The aim is to collect a large dataset of isolated Arabic words that covers all letters of the alphabet in all possible shapes using a small number of simple words. The end goal is to obtain a very large database of segmented letter images, which can be used to build and evaluate Arabic handwriting recognition systems that are based on segmented letter recognition. The current version of the dataset contains $25114$ samples of $109$ unique Arabic words that cover all possible shapes of all alphabet letters. The samples were collected from $907$ writers. In its current form, the dataset can be used for the problem of closed-vocabulary word recognition. We evaluated a number of window-based descriptors and classifiers on this task and obtained an accuracy of $92.16\%$ using a SIFT-based descriptor and ANN.
%Research in handwriting recognition is restrained by the availability of suitable datasets for evaluating the developed techniques. Due to the cursive nature of the Arabic scripting system, most techniques for open-vocabulary Arabic handwriting recognition rely on HMM to avoid segmenting words into characters. Other techniques require such segmentation, a process that is both challenging and difficult to evaluate due to the lack of ground truth data for it. We believe that a different approach has to be explored, which is handwriting recognition using simultaneous character detection and segmentation. To explore this direction, a large dataset is needed for Arabic alphabet letters that are extracted from complete handwritten words. In this paper, we present the first phase of AlexU-Word dataset, which is a step towards developing such large segmented-letter dataset. The current version of the dataset contains $25114$ unsegmented word images, covering $112$ different Arabic words, and collected from 907 different writers. In its current form, the dataset can be used for the problem of closed-vocabulary word recognition. We evaluated a number of window-based descriptors and classifiers on this task and obtained accuracy of over $90\%$ word-recognition rate.

\end{abstract}


\section{Introduction}
\label{sec:intro}
% different steps of Arabic OHR
The success of any handwriting recognition system is heavily dependent on the amount and nature of data used in training the system. Arabic handwriting recognition is not an exception. Over the past two decades, numerous datasets for Arabic handwriting recognition were collected to keep advancing the state of the art in the challenging task.

Techniques for Arabic handwritten word recognition can be classified into two main categories: word-based methods and character-based methods. In word-based methods, the segmentation of words into their constituting letters is not required. Typically, these methods rely on Hidden Markov Models (HMMs)~\cite{Poltz09} in the open-vocabulary scenario. In the closed vocabulary scenario, global features can be used with any classification algorithm, e.g. ~\cite{Alma'adeed:2006:ROH:1155446.1156039}. On the other hand, character-based methods require pre-segmentation of words into characters or parts of characters. After the segmentation, trained character models are used to classify groups of consecutive segments into characters~\cite{Mahmoud20141096}.

Coming from a computer vision background, we believe that a new paradigm for word recognition ought to be explored. This paradigm bases word recognition on individual character detection. Suppose that we can build strong models for recognizing individual characters in all possible shapes. These models can be deployed to find characters in a given input image without using heuristics for segmenting words into characters. In other words, word segmentation into characters can be obtained by detecting the presence and locations of the constituting characters of a word. To train such models, a very large collection of samples from alphabet characters, in all their possible shapes, has to be available. Towards this goal, we started collecting the AlexU-Word dataset.

The AlexU-Word dataset is collected with the main goal of having a large number of samples from few and simple words that contain all Arabic alphabet characters in all possible shapes. The current version of the dataset contains $25114$ samples from $109$ unique Arabic words. The future plan\footnote{Subject to availability of funding.} is to extend the dataset by adding more words and more samples, and to segment the collected words into their constituting characters in order to form a very large database of segmented characters.

The current version of the dataset is suitable for closed-vocabulary isolated word recognition. We evaluated a couple of global word descriptors with three different classifiers on this task. The best accuracy obtained was $92.16\%$ word recognition rate.

The rest of the paper is organized as follows. In Section~\ref{sec:relwork}, a summary of related work is presented. In Section~\ref{sec:alexuword}, details of the dataset collection and characteristics are presented. In Section~\ref{sec:exp}, our experimental evaluation on the word recognition task is presented. Finally, in Section~\ref{sec:conc}, the paper is concluded.

\section{Related Work}
\label{sec:relwork}
% other datasets and their limitations
Over the past $15$ years, many datasets have been collected for Arabic Offline Handwriting Recognition (AOHR). A comprehensive list can be found in a recent survey~\cite{parvez_offline_2013}. 

By far, the most commonly used and one of the largest publicly available datasets for AOHR is the IFN/ENIT dataset~\cite{Pechwitz02ifn/enit-}. It contains the names of over $900$ Tunisian cities, with over $26000$ samples in total. The dataset is useful for closed-vocabulary handwriting recognition. However, there is no effort made to uniformly cover all letters of the alphabet in all shapes, or to make it easy to segment the collected words into letters for segmented-letter training. 

Perhaps, the closest dataset to ours in the collection philosophy is ~\cite{Alamri08anovel}, where a small number of words were chosen to cover all possible shapes of Arabic alphabet characters. However, the number of word samples available is only $11375$, which is too far from our target. The current version of AlexU-Word dataset contains $25114$ samples. The closest dataset to ours in terms of the number of samples is Al-Isra' dataset~\cite{Kharma99}. It contains $37000$ word samples. However, we have not been able to find a way to obtain a copy from the dataset. There are other more specialized datasets. For example, \cite{Abdleazeem08} is only for numeral recognition. The dataset collected in ~\cite{AlMaadeedEH04} focused on the most commonly used $20$ words in Arabic, which may not have all variations of Arabic alphabet characters.

The KHATT dataset~\cite{Mahmoud20141096} is the most recent and largest dataset collected for handwritten Arabic. It contains close to $200,000$ samples covering a large collection of words from different topic areas in Arabic literature. The dataset may exhibit the natural distribution of characters in typical Arabic writing. However, it is still not designed with the task of character detection in mind.

\section{AlexU-Word Dataset}
\label{sec:alexuword}
AlexU-Word dataset contains $25114$ samples from $109$ different Arabic words. The word samples are collected from $907$ different writers. Among the writers, there were $662$ males and $245$ females; and $846$ right-handed and $61$ left-handed. Details of writer and word distribution are given in Table~\ref{tab:WriterWordDist}.
\begin{table}
	\centering
		\begin{tabular} {|c|c|c|c|}
		\hline
									& Male 					& Female 			& Total  \\ \hline
		Right-Handed  & $619|17149$ 	& $227|6301$ 	& $846|23450$ \\ \hline
		Left-Handed 	& $43|1180$ 		& $18|484$ 		& $61|1664$ \\ \hline
		Total 				& $662|18329$ 	& $245|6785$ 	& $907|25114$ \\ \hline
		\end{tabular}
	\caption{Writer and word distribution in AlexU-Word. Each cell contains two numbers. The first is the number of forms, and the second is the number of words. Due to exclusion of some words in the verification process, the number of words may not be an exact multiple of $28$ although each form contained exactly $28$ words.}
	\label{tab:WriterWordDist}
\end{table}

The $109$ Arabic words in the dataset were chosen to cover all possible cases for each letter in the alphabet. An Arabic word can be made of one or more connected groups of letters, each group is known as a Part of Arabic Word or a PAW. Each letter of the Arabic alphabet can have at most four different cases (hence, shapes), depending on its position with respect to the containing PAW: beginning of a PAW, middle of a PAW, end of a PAW, or isolated (i.e. a standalone PAW). Since the Arabic alphabet contains $28$ different letters, and each can have up to four different cases, at most $112$ words can cover all letters in all cases. We chose 109 unique words to collect in our dataset. The words were selected to be short and simple so that future segmentation of the words into letters becomes easy. In fact, the words were collected from an Android application that teaches children about the Arabic alphabet and their shapes\footnote{The app is called HOROOF LOCATION by Abo Mohannad. In Arabic, the app is called \<أشكال الحروف الهجائية ومواضعها> by \<ابومهند>.}. The chosen words are illustrated in the sample form models in Fig.~\ref{fig:sampleForms}.

\subsection{Data Collection}
\label{sec:dataCollection}
\begin{figure}
\centering
\subfloat[Model I]{\includegraphics[width=1.5in]{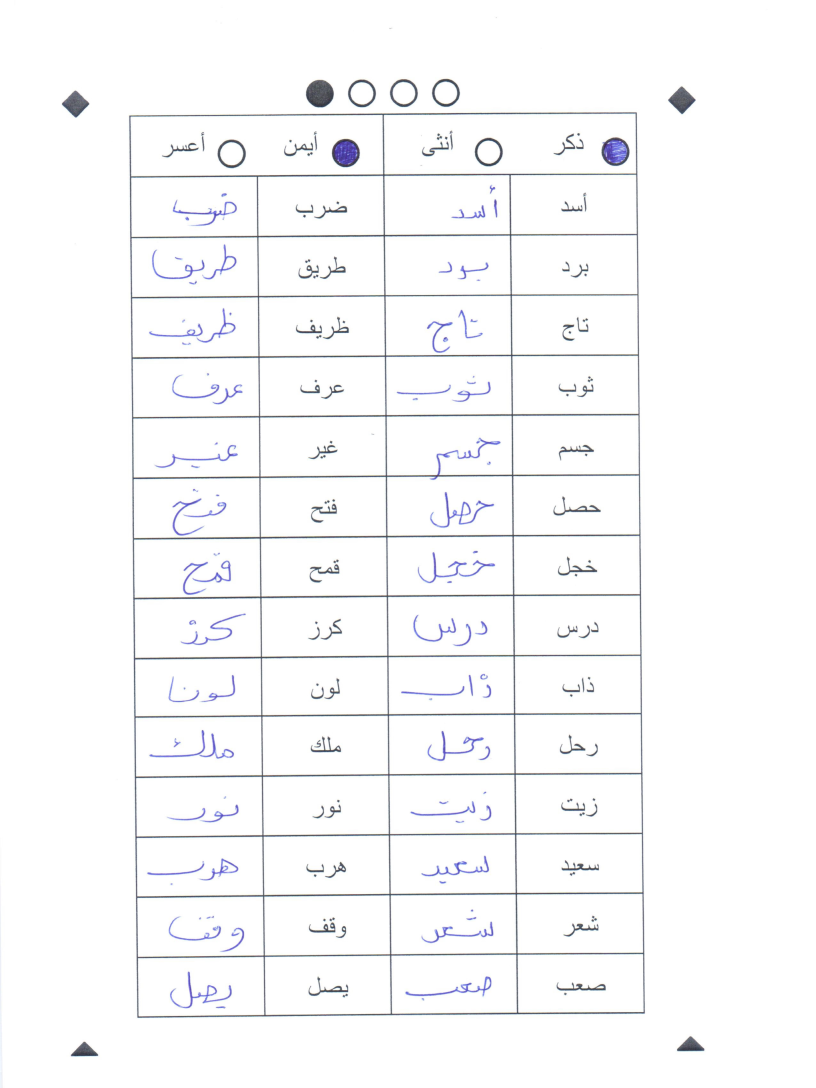}}%
\label{fig:sampleForms_model1}
\hfil
\subfloat[Model II]{\includegraphics[width=1.5in]{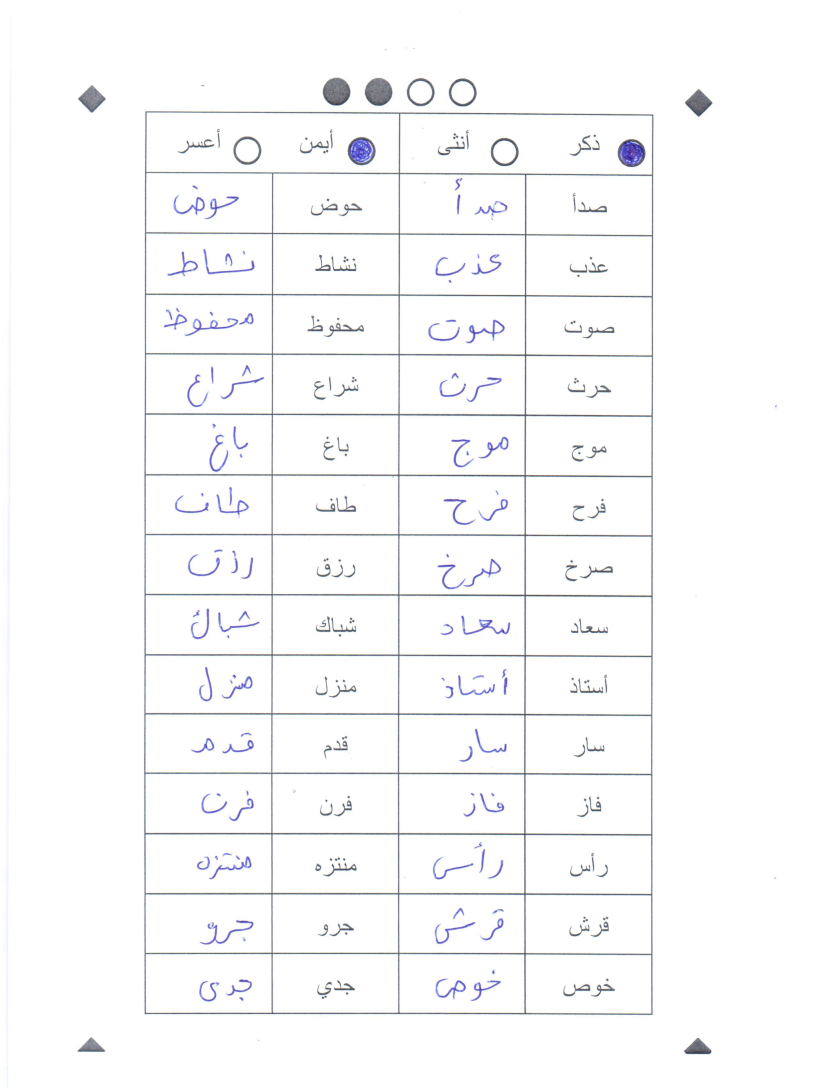}}%
\label{fig:sampleForms_model2}
\hfil
\subfloat[Model III]{\includegraphics[width=1.5in]{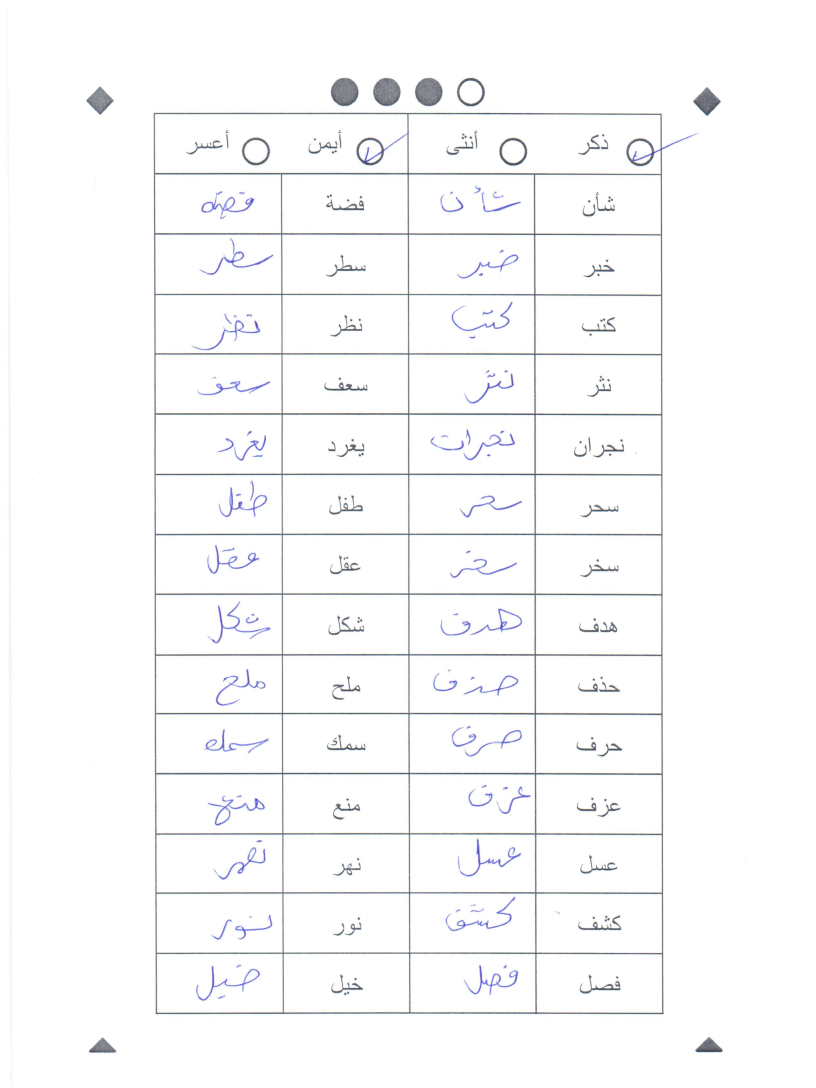}}%
\label{fig:sampleForms_model3}
\hfil
\subfloat[Model IV]{\includegraphics[width=1.5in]{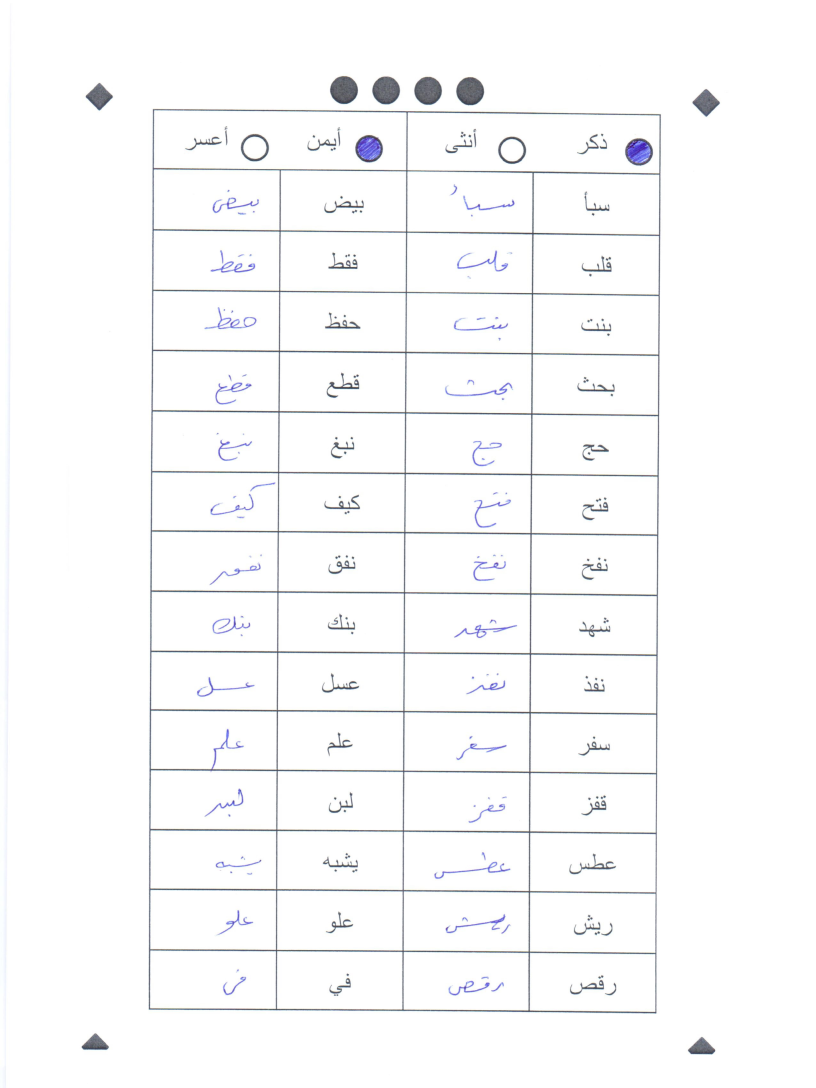}}%
\label{fig:sampleForms_model4}
\caption{Samples from the four form models used in collecting AlexU-Word dataset. The four models correspond to the four possible shapes of an Arabic letter in a word. Each model contains 28 words, each word contains one letter of the alphabet in the designated letter shape for the model. The number of filled circles at the top marks the form model. Beneath the form model number, two questions are put: one is about the gender of the writer, and one is about his/her handedness.}
\label{fig:sampleForms}
\end{figure}

AlexU-Word dataset was collected at the Faculty of Engineering, Alexandria University, Alexandria, Egypt. 928 different writers originally participated in the collection process; data from $14$ of them were discarded in the verification process to end up with data from $907$ writers only. All writers were in the ages from 18 to 25 years old. Most of the participants were students in the preparatory year of the faculty. Data was collected during the lab instructing times of the Introduction to Computers course. Each writer was asked to fill out only a one-page form. Due to the busy lab time, the form design was made as simple as possible to minimize the collection time. Each form contained a table of 28 Arabic words, where each word appeared printed in one cell of the table, next to it, an empty cell was left for the associated handwritten sample. Four different form models were used to cover all possible cases of Arabic letter, as explained above. A sample from each form appears in Fig.~\ref{fig:sampleForms}. Writers were provided with blue pens, all of the same kind, to facilitate form processing, as explained in Section~\ref{sec:processing}.
% The process of data collection was reported to take around $15$ minutes on average.

In addition to handwriting the form words, each writer also was asked to specify his/her gender and handedness. Distribution of writers among different form models is shown in Table~\ref{tab:WriterDistPerModel}. Distribution of word counts among different form models is shown in Table~\ref{tab:WordDistPerModel}. Note that although each form contains exactly $28$ words, the word counts are not always a multiple of $28$. This is due to the exclusion of some words in the verification process, as explained in Section~\ref{sec:processing}.

\begin{table}
	\centering
		\begin{tabular} {|c|c|c|c|c|}
		\hline
					Form Model		& 			I	& 			II	& 			III	& IV 				\\ \hline
		Male/Right-Handed  	& $143$ 	& $233$ 		& $150$ 		& $93$ 			\\ \hline
		Male/Left-Handed 		& $14$ 		& $13$ 			& $9$ 			& $7$  			\\ \hline
		Female/Right-Handed & $55$ 		& $64$ 			& $78$ 			& $30$ 			\\ \hline
		Female/Left-Handed 	& $2$ 		& $5$ 			& $6$ 			& $5$ 			\\ \hline
		Total 							& $214$ 	& $315$ 		& $243$ 		& $135$ 		\\ \hline
		\end{tabular}
	\caption{Writer distribution among form models and writer information.}
	\label{tab:WriterDistPerModel}
\end{table}

\begin{table}
	\centering
		\begin{tabular} {|c|c|c|c|c|c|}
		\hline
					Form Model		& 			I	& 			II	& 			III	& IV 		\\ \hline
		Male/Right-Handed  	& $3983$ 	& $6492$ 		& $4132$ 		& $2542$\\ \hline
		Male/Left-Handed 		& $386$ 	& $364$ 		& $248$ 		& $182$ \\ \hline
		Female/Right-Handed & $1536$ 	& $1788$ 		& $2150$ 		& $827$ \\ \hline
		Female/Left-Handed 	& $56$ 		& $124$ 		& $168$ 		& $136$ \\ \hline
		Total 							& $5961$ 	& $8768$ 		& $6698$ 		& $3687$ \\ \hline
		\end{tabular}
	\caption{Word distribution among form models and writer information. Due to the exclusion of some words in the verification process, the number of words may not be an exact multiple of $28$ although each form contained exactly $28$ words.}
	\label{tab:WordDistPerModel}
\end{table}

\subsection{Form Processing and Verification}
\label{sec:processing}
\begin{figure}
\centering
\subfloat[Scanned Form]{\includegraphics[height=2in]{sampleFormModel1.png}}%
\label{fig:sampleForms_model1_scanned}
\hfil
\subfloat[Transformed to Canonical Form]{\includegraphics[height=2in]{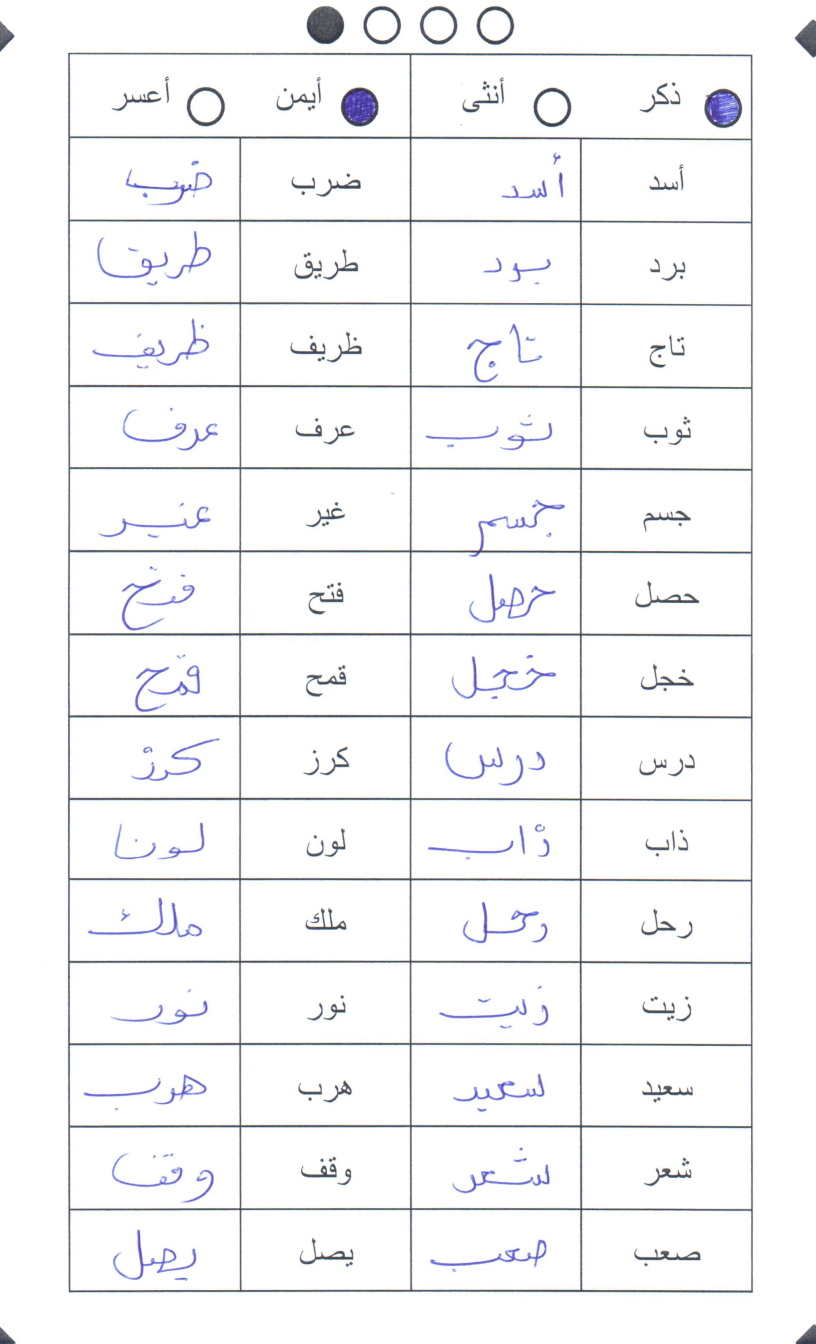}}%
\label{fig:sampleForms_model1_aligned}
\caption{The scanned form in (a) is transformed to the canonical upright pose and size in (b) for automatic extraction of handwritten words and writer's information. The four rhombus shaped markers at the four corners are detected and the transformation is performed according to their positions.}
\label{fig:formAlignment}
\end{figure}

Each collected form was scanned at the resolution of $600$ dpi. This gives us forms similar to the ones shown in Fig.~\ref{fig:sampleForms}. As can be observed in the figure, scanned forms can be slightly rotated or translated with respect to a perfectly upright position. To automatically extract handwritten samples, forms are first transformed to a canonical size and upright position. To achieve this, the rhombus-shaped markers at the four corners of the form are detected, and the form is transformed according to their positions to the upright canonical pose, as shown in Fig.~\ref{fig:formAlignment}. The markers were detected using a version of the form that is scaled $33$ times down. On this version, Harris corner detection was applied and the four outermost detected corners were considered corresponding to the four rhombus-shaped markers.

After the form is transformed to the canonical form, the location of writer information fields and handwritten words become known. Since writers were asked to use blue pens, only the blue channel of the canonical form was processed. The blue channel is binarized to give only the ink areas. By comparing the overall ink-stained areas in writer info's options, we can determine which choice he/she made, regardless of the way the writer used to highlight them, e.g. a check mark or mere filling. Binary images for handwritten words were extracted from each form, empty areas surrounding a word were removed so that the stored binary image contains only the word without any margin.

\begin{figure}
\centering
\subfloat[]{\includegraphics[height=1.25in]{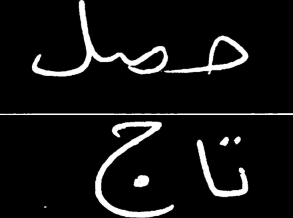}}%
\label{fig:sampleLoose}
\hfil
\subfloat[]{\includegraphics[height=1.25in]{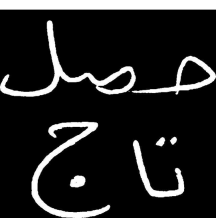}}%
\label{fig:sampleCorrected}
\caption{Sample of binary word images that are loosely cropped in (a) and the corrected version after verification in (b).}
\label{fig:wordCorrection}
\end{figure}

\begin{figure}
\centering
\subfloat[]{\includegraphics[width=1in]{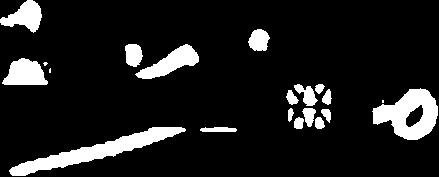}}%
\label{fig:sampleBadScan}
\hfil
\subfloat[]{\includegraphics[width=1in]{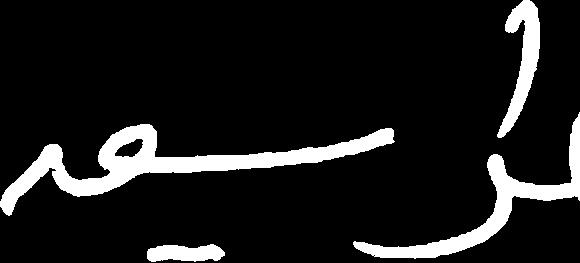}}%
\label{fig:samplePartOfWord}
\hfil
\subfloat[]{\includegraphics[width=0.9in]{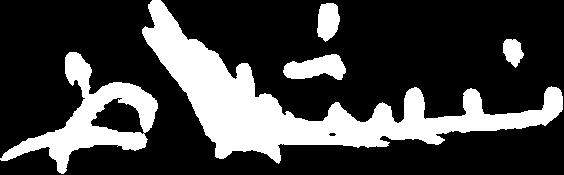}}%
\label{fig:sampleScratch}
\hfil
\subfloat[]{\includegraphics[width=0.1in]{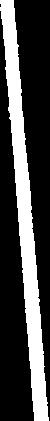}}%
\label{fig:sampleBG}
\caption{Samples of binary word images that were removed after manual verification due to scanner errors (a), word too large for the cell and hence truncated (b), word containing too much scratch (c), or background area cropped instead of word (d).}
\label{fig:wordRemoval}
\end{figure}

Due to errors in scanning and due to the possibility of imperfect parameter adjustments in the previous steps, some extracted word images either were not tight around the word (i.e. contained some nearby noise spots), contained only a part of the word, contained only a part of the background, or unrecognizable handwriting due to severe scanning errors or writer's scratching. All extracted images were manually inspected by the authors. In the case of images that are loosely cropped, extra margins were removed. In the other cases, the images were completely removed from the dataset. Examples of cropping corrections are shown in Fig.~\ref{fig:wordCorrection}. Examples of removed images are shown in Fig.~\ref{fig:wordRemoval}.

\subsection{Download}

The dataset can be downloaded through this link: \url{http://www.eng.alexu.edu.eg/~mehussein/alexu-word/}.
\section{Experiments}
\label{sec:exp}
The ultimate goal after collecting AlexU-Word dataset is to perform letter segmentation and train classifiers on individual characters. Until the whole dataset is collected and letter segmentation is performed, the dataset can be used for evaluating closed-vocabulary word-recognition methods. In this section, we provide base-line comparison among a couple of window-based descriptors, inspired from the computer-vision literature, across a number of classification algorithms. In Section~\ref{sec:feats_cls}, our window-based descriptors are explained. In Section~\ref{sec:expSetup}, our experimental setup is explained. In Section~\ref{sec:wordrec}, the word classification results are presented.

\subsection{Window-Based Descriptors}
\label{sec:feats_cls}
In another study of our group, window-based descriptors were introduced and compared on the task of isolated Arabic letter recognition~\cite{Torki14}. Two of the top performing descriptors in this work were based on the Histograms of Oriented Gradients (HOG) descriptor~\cite{Dalal05} and the Scale Invariant Feature Transform (SIFT) descriptor~\cite{Lowe04}, two very popular descriptors in the object detection and recognition literature.

\begin{figure}
	\centering
		\includegraphics[width=3in]{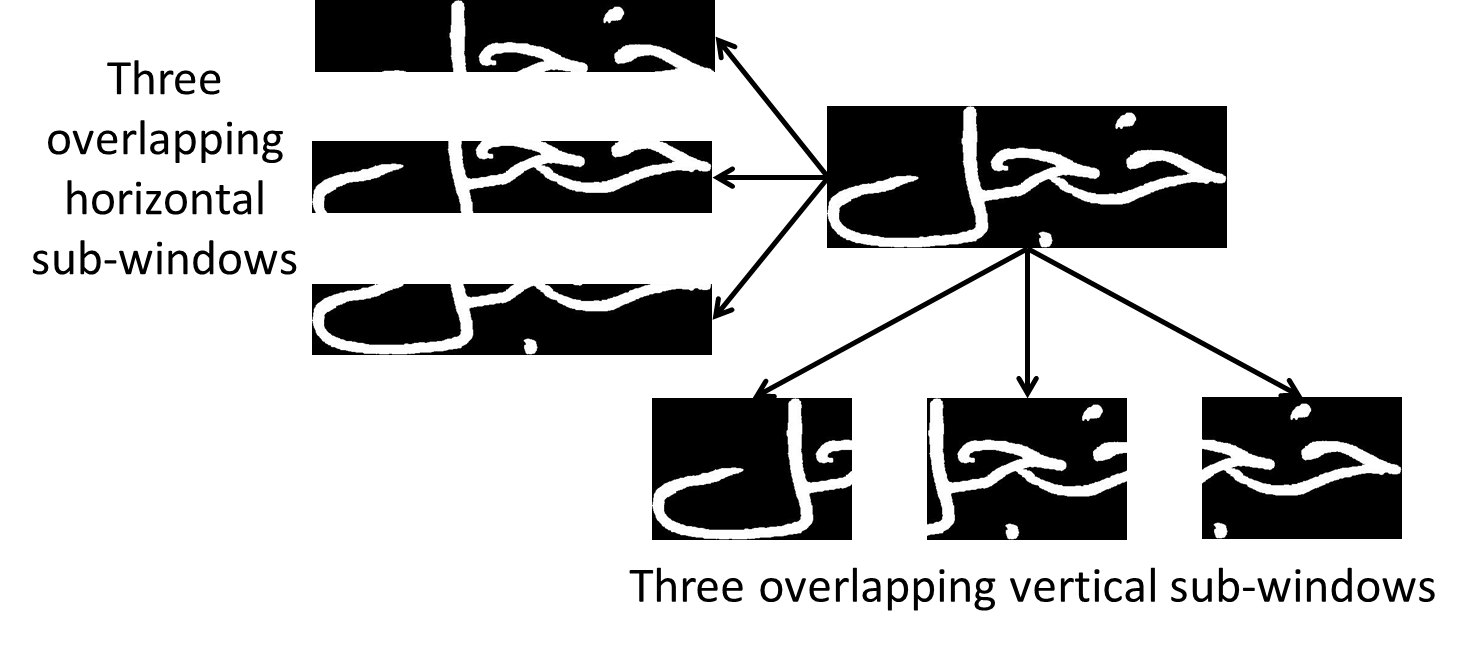}
	\caption{Construction of window-based descriptor using the whole image and 6 overlapping sub-windows.}
	\label{fig:overlappingSubwindows}
\end{figure}

In order to better represent the shape of a letter, Torki et al.~\cite{Torki14}, used a spatial pyramid construction of the HOG and SIFT descriptor with overlapping sub-windows, which they called HOG7 and SIFT7. The number $7$ at the end of each descriptor's name represents the number of sub-windows of the input image on which the descriptor is constructed. Particularly, the descriptor is computed as follows. The base descriptor (HOG or SIFT) is constructed once for the entire image. Then, it is constructed three times for three overlapping vertical strips from the image, each of which has half the image's width with an equal step between each consecutive two of them. Finally, similarly, it is constructed three times for three overlapping horizontal strips from the image, each of which has half the image's height with an equal step between each consecutive two of them. Illustration of the overlapping sub-window division is shown in Figure~\ref{fig:overlappingSubwindows}. The descriptors from the seven windows are concatenated to make the final descriptor.

\subsection{Experimental Setup}
\label{sec:expSetup}
The two descriptors, HOG7 and SIFT7 (Section~\ref{sec:feats_cls}), are experimented with three classification algorithms, k-Nearest Neighbor (k-NN), Artificial Neural Networks (ANN), and Support Vector Machines (SVM). To train each classifier, the word samples in the dataset are divided into training (around $60\%$), testing (around $20\%$), and validation sets (around $20\%$). Details of division of words over the three sets are given in Table~\ref{tab:evaluationSetup}.

\begin{table}
	\centering
		\begin{tabular} {|c|c|c|c|}
		\hline
								& Training 	& Testing 	& Validation \\ \hline
		Model I			& $3588$ 		& $1184$ 		& $1189$ 		\\ \hline
		Model II 		& $5259$ 		& $1763$ 		& $1746$ 		\\ \hline
		Model III  	& $3980$ 		& $1368$ 		& $1350$ 		\\ \hline
		Model IV 		& $2212$ 		& $733$ 		& $742$ 		\\ \hline
		Total 			& $15039$ 	& $5048$ 		& $5027$ 		\\ \hline
		\end{tabular}
	\caption{Division of dataset words from each form model over the training, testing, and validation sets.}
	\label{tab:evaluationSetup}
\end{table}

Initially, the training set and evaluation on the validation set was done with different values of classifiers' parameters until the best set of parameters was determined. Then, the training and validation sets were combined to train the classifier using the best parameters. The final reported accuracy is based on applying this classifier on the testing set. The best parameters for each classifier-descriptor pair are shown in Table~\ref{tab:classifierParams}. Note that for ANN, one hidden layer of $250$ nodes was always used. Also, the activation function was the sigmoid function, the number of iterations in this validation phase was fixed at $250$ iterations, and the optimization algorithm used was gradient descent. For SVM, only the linear kernel was used.

\begin{table}
	\centering
		\begin{tabular} {|c|c|c|}
		\hline
					& HOG7						& SIFT7 				\\ \hline
		k-NN	& $k=9$ 					& $k=5$ 				\\ \hline
		ANN 	& $\lambda=0.001$ & $\lambda=0.03$	\\ \hline
		SVM  	& $C=0.3$ 					& $C=3$ 				\\ \hline
		\end{tabular}
	\caption{Best classifier parameters obtained after training on the training set and evaluation on the validation set. $\lambda$ is the regularization coefficient for ANN.}
	\label{tab:classifierParams}
\end{table}

\subsection{Word Recognition Results}
\label{sec:wordrec}
The final classification accuracies for the three classifiers are shown in Figure~\ref{tab:wordRecResults}. These results are obtained by training each classifier on the combination of training and validation sets using the best parameters obtained on the validation set, as shown in Table~\ref{tab:classifierParams}. The other fixed parameters, as explained in Section~\ref{sec:expSetup}, remained the same here, except for the number of iterations in ANN, which was raised to $400$ in this phase.

\begin{table}
	\centering
		\begin{tabular} {|c|c|c|}
		\hline
					& HOG7			& SIFT7 		\\ \hline
		k-NN	& $67.45\%$ & $79.73\%$ \\ \hline
		ANN 	& $86.67\%$ & $92.16\%$	\\ \hline
		SVM  	& $86.13\%$ & $91.16\%$ \\ \hline
		\end{tabular}
	\caption{Classification accuracy based on training each classifier on the combination of training and validation sets using the parameters in Table~\ref{tab:classifierParams}, and evaluation on the testing set.}
	\label{tab:wordRecResults}
\end{table}

The superiority of the SIFT7 descriptor over the HOG7 descriptor with all classifiers is clear. Even the naive k-NN classifier performs quite well with SIFT7 by scoring $80.59\%$ accuracy, with a big difference from HOG7, which scores $65.47\%$ accuracy with the same classifier.

In terms of classifiers, k-NN is understandably the least performing, and ANN is slightly superior to SVM. Superiority of ANN can be due to its non-linearity since we only used the linear kernel with SVM.

\begin{figure}
	\centering
		\includegraphics[width=3.5in]{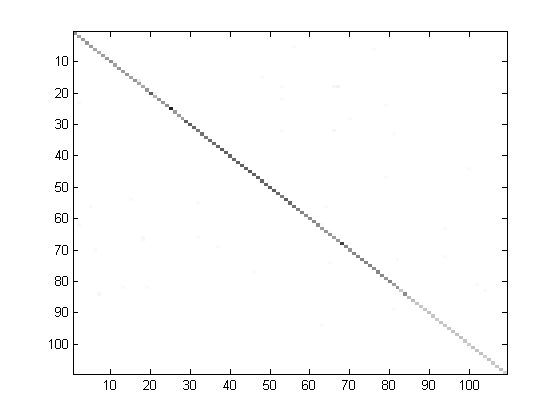}
	\caption{Confusion matrix for the 109 classes on the best performing descriptor-classifier pair, SIFT7-ANN. Black means 1 and white means 0.}
	\label{fig:confMat}
\end{figure}

\begin{figure}
	\centering
		\includegraphics[width=1.2in]{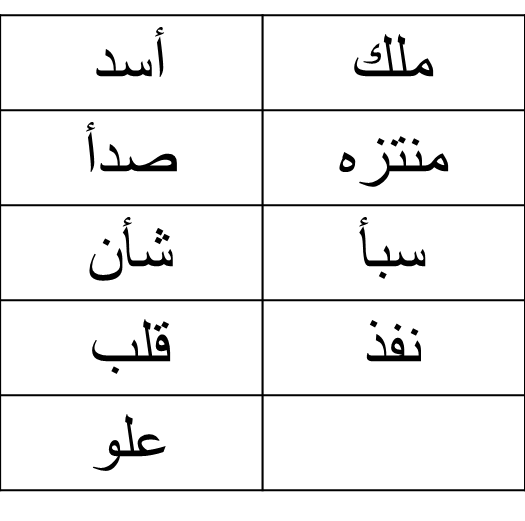}
	\caption{The nine classes that can be recognized with $100\%$ accuracy by the best descriptor-classifier combination.}
	\label{fig:leastConfusing}
\end{figure}

\begin{figure}
	\centering
		\includegraphics[width=1.2in]{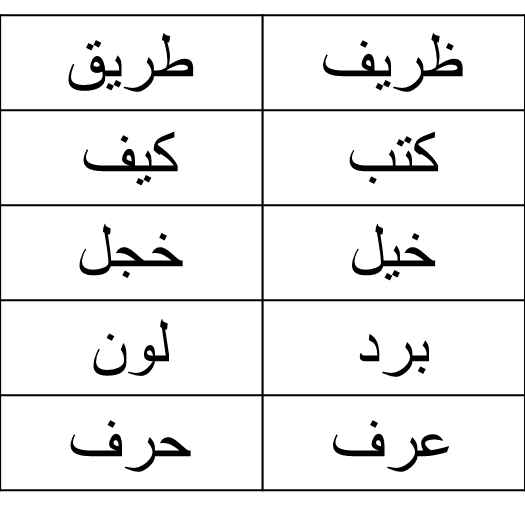}
	\caption{Most confusing five pairs of words in descending order of confusion.}
	\label{fig:mostConfusing}
\end{figure}

The confusion matrix of the best classifier is shown in Figure~\ref{fig:confMat}. The confusion matrix is strongly sparse and diagonal, which indicates that the classifier can easily distinguish between most pairs of classes. In fact, there are nine word classes that can be classified with $100\%$ accuracy, which are shown in Figure~\ref{fig:leastConfusing}. Inspecting the confusion matrix for the most confusing classes, the most confusing five pairs of words were determined to be the ones shown in Figure~\ref{fig:mostConfusing}. It is clear that these pairs indeed include very similar words.
% 17-16, 104-59, 7-84, 2-23, 18-66

\section{Conclusion}
\label{sec:conc}
We have introduced the AlexU-Word dataset, which includes $25114$ samples of $109$ unique Arabic words. The words are chosen to be few in number, simple in nature, and covering all possible shapes of all Arabic alphabet characters. The future plan is to extend this dataset and segment its words into characters for the purpose of building very accurate models for character recognition and localization. We believe that availability of such data can change the way offline Arabic handwriting recognition is done. We presented experimental evaluation on the collected data for the task of closed-vocabulary, isolated word recognition. Our best descriptor-classifier combination obtained $92.16\%$ correct recognition rate.

\section*{Acknowledgment}
The authors would like to thank the Center of Excellence for Smart Critical Infrastructure (SmartCI) at Virginia Tech - Middle East and North Africa (VT-MENA) for sponsoring this work. A special thanks is due to Dr. Yosry Taha, the coordinator of the preparatory year course on Introduction to Computer at the Faculty of Engineering, Alexandria University, for giving us the permission to perform data collection from students during the course's lab instructing times. Finally, thanks are due to all the teaching assistants of the course who helped in data collection.



\bibliographystyle{IEEEbib}
%\bibliography{refs}

\end{document}